\documentclass[10pt,twocolumn,letterpaper]{article}

\usepackage{iccv}
\usepackage{times}
\usepackage{graphicx}
\usepackage{epstopdf}
\usepackage{amsmath}
\usepackage{amssymb}

\usepackage{bm}
\usepackage{enumitem}
\usepackage{booktabs}
\usepackage{multirow}
\usepackage{array}
\usepackage{float}
\usepackage{xcolor}
\usepackage[pagebackref=true,breaklinks=true,letterpaper=true,colorlinks,bookmarks=false]{hyperref}

\iccvfinalcopy 

\begin{document}

\title{Edge-Direct Visual Odometry}

\author{Kevin Christensen \\
The Robotics Institute\\
Carnegie Mellon University\\
{\tt\small kchrist1@andrew.cmu.edu }
\and
Martial Hebert\\
The Robotics Institute\\
Carnegie Mellon University\\
{\tt\small mhebert@andrew.cmu.edu}
}

\maketitle

\begin{abstract}
   In this paper we propose an edge-direct visual odometry algorithm that efficiently utilizes edge pixels to find the relative pose that minimizes the photometric error between images.  Prior work on exploiting edge pixels instead treats edges as features and employ various techniques to match edge lines or pixels, which adds unnecessary complexity.  Direct methods typically operate on all pixel intensities, which proves to be highly redundant.  In contrast our method builds on direct visual odometry methods naturally with minimal added computation. It is not only more efficient than direct dense methods since we iterate with a fraction of the pixels, but also more accurate.  We achieve high accuracy and efficiency by extracting edges from only one image, and utilize robust Gauss-Newton to minimize the photometric error of these edge pixels.  This simultaneously finds the edge pixels in the reference image, as well as the relative camera pose that minimizes the photometric error. We test various edge detectors, including learned edges, and determine that the optimal edge detector for this method is the Canny edge detection algorithm using automatic thresholding.  We highlight key differences between our edge direct method and direct dense methods, in particular how higher levels of image pyramids can lead to significant aliasing effects and result in incorrect solution convergence.  We show experimentally that reducing the photometric error of edge pixels also reduces the photometric error of all pixels, and we show through an ablation study the increase in accuracy obtained by optimizing edge pixels only.  We evaluate our method on the RGB-D TUM benchmark on which we achieve state-of-the-art performance.
\end{abstract}

\section{Introduction}

Visual odometry (VO), or the task of tracking camera pose from a stream of images, has received increased attention due to its widespread applications in robotics and augmented reality.  Camera tracking in unknown environments is one of the most difficult challenges of computer vision.  

While VO has become a more popular area of research, there are still several challenges present.  Such challenges are operating in low-texture environments, achieving higher frame rate processing capabilities for increased positional control, and reducing the drift of the trajectory estimate.  Any new algorithm must also deal with inherent challenges of tracking camera pose, in particular they must be able to handle the high bandwidth image streams, which requires efficient solutions to extract useful information from such large amounts of data.

\begin{figure}[t]
\begin{center}
\includegraphics[width=1.0\linewidth]{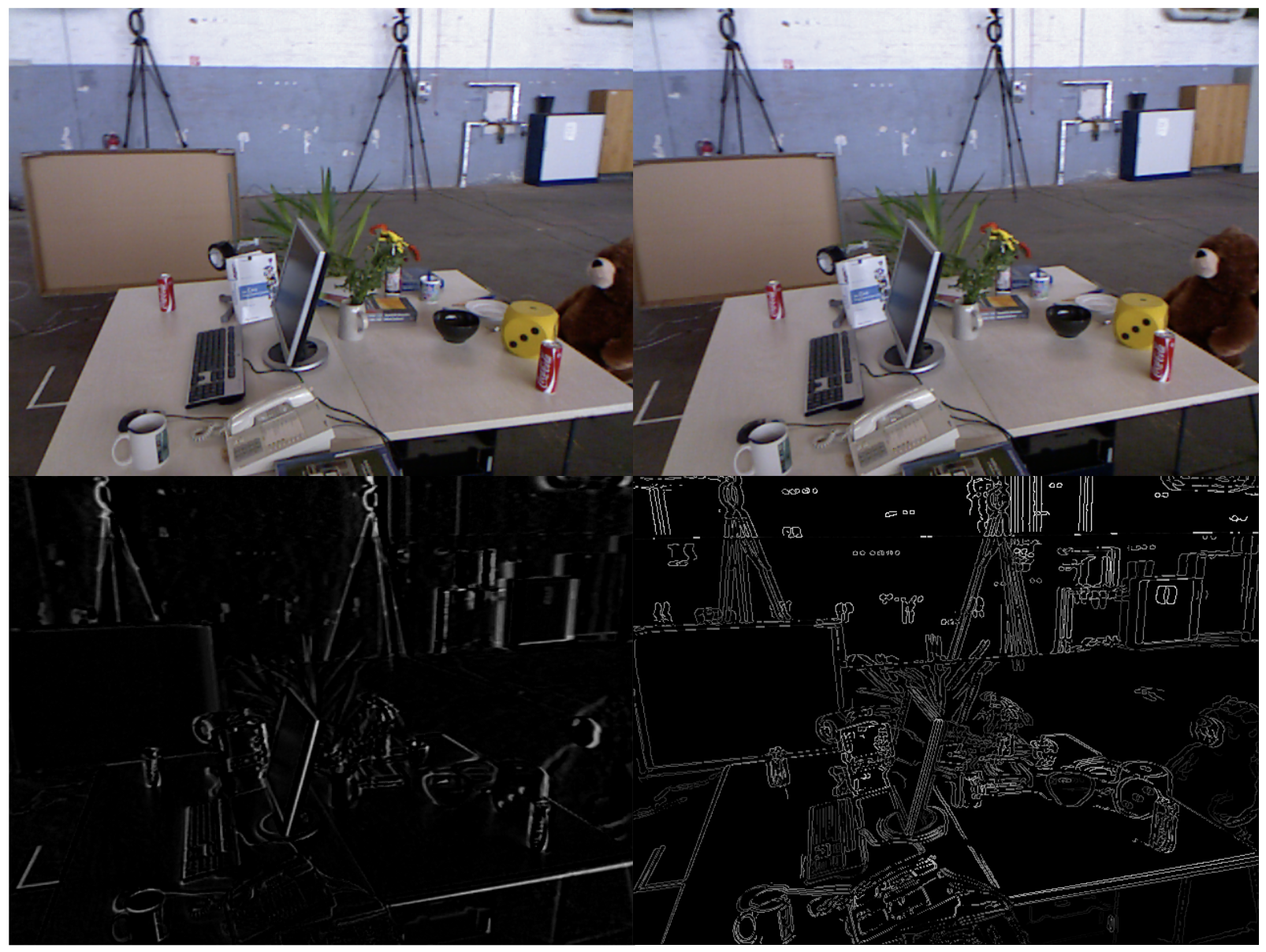}
\end{center}
   \caption{In contrast to previous direct methods that attempt to minimize the photometric error (bottom left) between reference frame (top left) and input image (top right), we minimize the photometric error of only the edges (bottom right).}
\label{fig:beg}
\end{figure}

\subsection{Contributions}
In this paper we propose a sparse visual odometry algorithm that efficiently utilizes edges to track the camera motion with state-of-the-art accuracy quantified by low relative pose drift.  More formally, we outline our main contributions:
\begin{itemize}
    \item An edge-direct visual odometry algorithm that outperforms state-of-the-art methods in public datasets.
    \item We provide experimental evidence that edges are the essential pixels in direct methods through an ablation study.
    \item We compare our edge method relative to a direct dense method. 
    \item We present key differences on reducing photometric error on edges as opposed to full image intensities.
    \item We optimize our algorithm with respect to several different types of edges.
\end{itemize}


\subsection{Visual Odometry vs. SLAM}

Simultaneous localization and mapping (SLAM) algorithms have taken visual odometry algorithms a step further by jointly mapping the environment, and performing optimization over the joint poses and map.  Additionally, SLAM algorithms implement loop closure, which enables systems to identify locations which it has visited before and optimize the trajectory by matching feature points against the prior image in memory.  

With the success of Bundle Adjustment and loop closure in producing near drift-free results, much of the attention has shifted from the performance of visual odometry algorithms to overall system performance. In reality the two are tightly coupled, and it is very important that visual odometry provides low-drift pose for two reasons.  Firstly, Bundle Adjustment requires a good initialization in order for it to converge to a drift-free solution.  Secondly, it is computationally expensive and is comparatively slow compared to the high frame-rate at which visual odometry performs.  For these reasons we focus solely on VO performance in this work, and we show competitive performance even against such SLAM systems.

\section{Related Work}
There are several different formulations of SLAM and VO algorithms.  Consequently, these algorithms can be classified as either \textit{direct} or \textit{indirect}, and as \textit{dense} or \textit{sparse}.

Indirect methods, also commonly referred to as feature-based methods, extract and match features.  These features are engineered representations of high-level features in the image, usually corners.  They must also store a representation of the feature that can be matched at a later frame.  This introduces two large sources of error.  After matching features, such methods calculate the fundamental/essential matrix or homography for planar scenes and scenes with low parallax.  

Due to the high level of inaccuracies present in feature extraction and matching, such algorithms must compute the fundamental matrix or homography in a RANSAC loop.  While feature-based methods have achieved accurate results, they remain computationally wasteful due to their reliance on RANSAC for robust estimation of such parameters.  Several examples of such systems that use indirect methods are ORB-SLAM, ORB-SLAM2~\cite{orbslam,orbslam2} and Parallel Tracking and Mapping (PTAM)~\cite{ptam}. Alternatively, direct methods directly use the sensor inputs, such as image intensities, to optimize an error function to determine relative camera pose.  

In addition to being classified as direct or indirect, SLAM and VO algorithms can additionally be classified as dense or sparse.  Dense methods have the advantage that they use all available information in the image, and can generate dense maps which is useful for robot navigation, for example.  Sparse methods have the advantage that since there are less points, it is generally less computationally expensive which can lead to large computational savings, especially if the algorithm requires many iterations or a loop to converge to a solution.

There have been many iterations of direct dense methods such as direct dense VO in~\cite{cremersrgbd}, RGB-D SLAM~\cite{densergbdslam}, and LSD-SLAM~\cite{engel14eccv}.  Even using dense methods, these systems achieve real-time performance on modern CPUs due to the highly efficient nature of these types of algorithms.  More recent advances highlight the fact that the information contained in image intensities are highly redundant, and attempt to minimize the photometric error only over sparse random points in the image in order to increase efficiency and thus speed \cite{DSO}.  Another direct method that has been used with success is the iterative closest point (ICP) algorithm, which is used in systems such as~\cite{kinect, kaesskintinuous}.  These systems minimize the difference between point alignment in contrast to image intensities.

The extension of direct methods using edge pixels is a logical direction, yet to the best of our knowledge no work has solely used edge pixels in a direct method minimizing the photometric error. In~\cite{robustedges} the authors reduce a Euclidean geometric error using the distance transform on edges which does not utilize all information available in the scene.  In~\cite{edgeenhanced} the authors minimize a joint error function combining photometric error over all pixels along with geometric error over edge pixels.  Minimizing a joint error function always suffers from the decision on how best to weight each function, and the weighting can have significant effect on the final converged solution.  In~\cite{DSO}, the authors threshold by gradients, which does not guarantee edges due to noise. They additionally select texture-less regions as well. 

We hypothesize direct methods have often avoided solely using edges due to several pitfalls when extending direct methods.  In particular, there is the question of which edges to use among the reference and the new image.  We have found that selecting the wrong edges produces incorrect convergence.  Additionally, edges are inherently unstable as changes in intensity and geometric position result in large changes in intensity and geometric position respectively, and incorrect formulation of the problem results in an incorrect solution.

\begin{figure}[t]
\begin{center}
\includegraphics[width=1.0\linewidth]{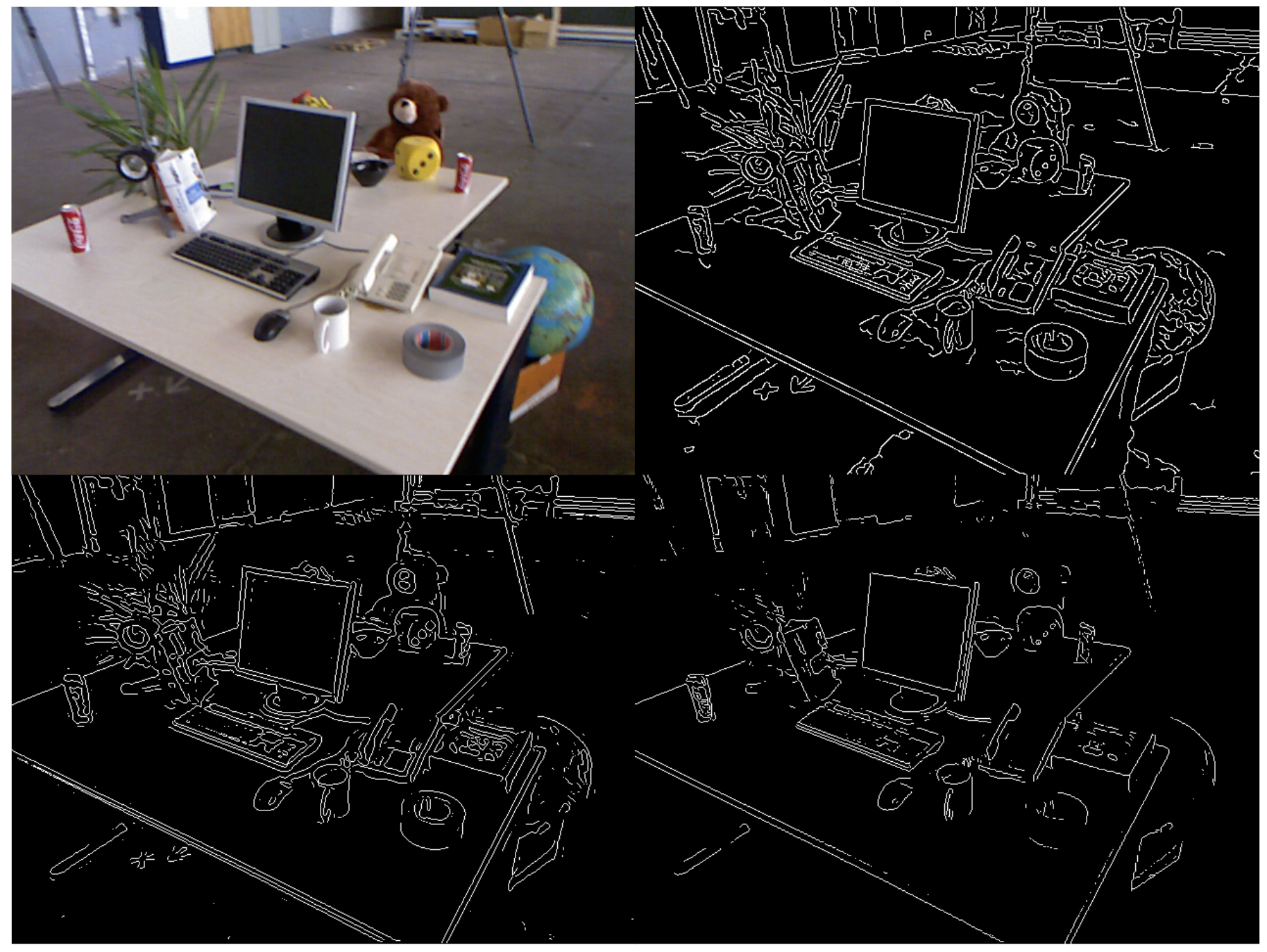}
\end{center}
   \caption{Examples of various edge extractions.  Top left: Original. Top right: Canny edges.  Bottom left: LoG edges.  Bottom right: Sobel edges.}
\label{fig:edgecomp}
\end{figure}

There have also been several indirect systems that have experimented with various strategies for utilizing edge information to track camera pose~\cite{Eade:edgels, klein:murray:eccv2008, edgeslam1, realtimeedgebased}.  Such methods treat edges as features and use complicated matching strategies which increase computation and add unnecessary heuristics.  In contrast our method simply extracts edges and incorporates it in a direct method.  This simple yet elegant and highly efficient sparse direct method provides lower drift than previous state-of-the-art visual odometry methods.  

Any system that extracts edges must choose between several edge extraction algorithms.  The most prominent are Canny edges \cite{canny}, followed by edges extracted from Laplacian of Gaussian (LoG) filters which are efficiently implemented using Difference of Gaussians (DoG).  Another type of edge that is not as popular but is very simple are Sobel edges.  More recently, there has been research involving the learning of edge features.  In \cite{Dollar} the authors utilize structured forests, and in \cite{HED} the authors utilize deep learning.  Instead of selecting one, we test various edge extraction algorithms with our system select the optimal edge extraction algorithm.  Note that \cite{HED} requires the use of a GPU and is far from real-time, so we do not consider this method.

\begin{figure*}[t]
\begin{center}
\includegraphics[width=1.0\linewidth]{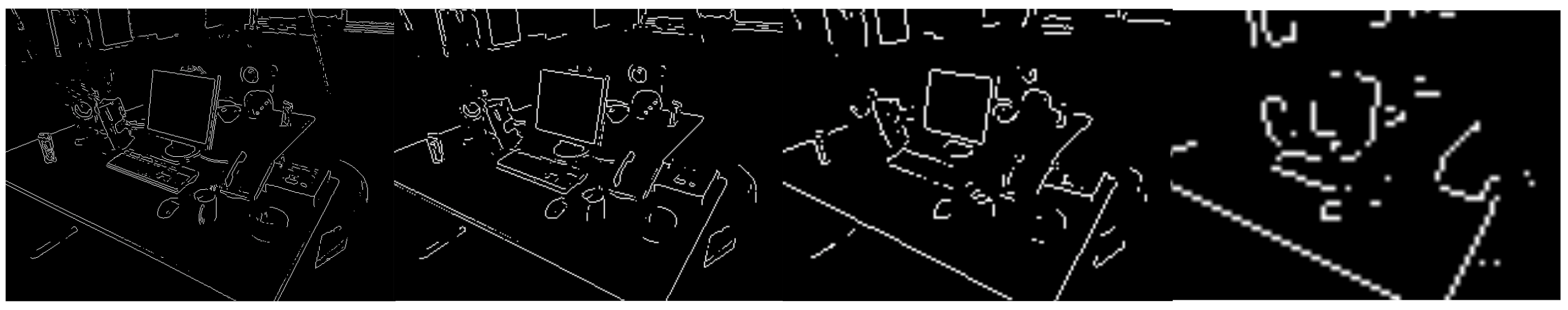}
\end{center}
   \caption{Edge pyramid for Canny edges. From left to right: Image 1: 640$\times$480, Image 2: 320$\times$240, Image 3: 160$\times$120, Image 4: 80$\times$60.   Any pyramid greater than three edge images deep starts to suffer from heavy amounts of aliasing, which led us to cut off our edge pyramid at the third level.}
\label{fig:aliasing}
\end{figure*}




\section{Edge-Direct Visual Odometry}
\subsection{Overview}
In this section we formulate the edge direct visual odometry algorithm.  The key concept behind direct visual odometry is to align images with respect to pose parameters using gradients.  This is an extension of the Lucas-Kanade algorithm~\cite{LKframe, Lucas:Kanade}.  

At each timestamp we have a reference RGB image and a depth image.  When we obtain a new frame, we assume we only receive an RGB image.  This enables our method to be extended to monocular VO by simply keeping a depth map and updating at each new time step.  Note also that we convert the RGB image into a grayscale image.

The key step of our algorithm is that we then extract edges from the \textit{new} image, and use them as a mask on the \textit{reference} image we are localizing with respect to.  We then align the images by iteratively minimizing the photometric error over these edge pixels.
The objective is to minimize the nonlinear photometric error

\begin{equation}\label{eq:1}
    r_i(\bm{\xi}) = \mathcal{I}_2(\tau ( \bm{x_i}, d_i, \bm{\xi})) - \mathcal{I}_1(\bm{x_i}), 
\end{equation}

where $\tau$ is the warp function that maps image intensities in the second image to image intensities in the first image through a rigid body transform.  
The warping function $\tau ( \bm{x_i}, d_i, \bm{\xi})$  is dependent on the pixel positions $\bm{x_i}$, the depth $d_i$ of the corresponding 3D point, and the camera pose $\bm{\xi}$.  Note that now the pixels we are using are only edge pixels, ie.  \begin{equation}\label{eq:2}
    \bm{x_i} \in \mathcal{E}(\mathcal{I}_2),
\end{equation}

where $\mathcal{E}(\mathcal{I}_2)$ are the edges of the new image.

\subsection{Camera Model}
In order to minimize the photometric error we need to be able to associate image pixels with 3D points in space.  Using the standard pinhole camera model, which maps 3D points to image pixels, we have 
\begin{equation}\label{eq:3}
    \pi(\bm{P}) = \begin{pmatrix} \frac{f_x X}{Z} + c_x , & \frac{f_y Y}{Z} + c_y  \end{pmatrix}^T ,
\end{equation}

where $f_x$ and $f_y$ are the focal lengths and $c_x$ and $c_y$ are the image coordinates of the principal point.  If we know the depth then we can find the inverse mapping that takes image coordinates and backprojects them to a 3D point $\bm{P}$ in homogenous coordinates

\begin{equation}\label{eq:4}
    \bm{P} = \pi^{-1}(\bm{x}_i, Z) =  \begin{pmatrix} \frac{x - c_x}{f_x}Z , & \frac{y - c_y}{f_y}Z , & Z, 1 \end{pmatrix}^T .
\end{equation}

\subsection{Camera Motion}

We are interested in determining the motion of the camera from a sequence of frames, which we model as a rigid body transformation.  The camera motion will therefore be in the Special Euclidean Group $SE(3)$.  The rigid body transform is given by $\bm{T} \in SE(3)$

\begin{equation}\label{eq:5}
    \bm{T} = 
    \begin{bmatrix} \bm{R} & \bm{t} \\
                    \bm{0} & 1 \\
    \end{bmatrix},
\end{equation}

where $\bm{R}$ is a $3\times3$ rotation matrix and $\bm{t}$ is a $3\times1$ translation vector.
Since we are performing Gauss-Newton optimization, we need to parameterize camera pose as a 6-vector through the exponential map $\bm{T} = exp_{\mathfrak{se}(3)}(\bm{\xi})$ so that we can optimize over the $SO(3)$ manifold for rotations.  At each iteration we can compose the relative pose update $\Delta \xi$ with the previous iteration estimate.

\begin{equation}\label{eq:6}
    \bm{\xi}^{(n+1)} = \Delta \bm{\xi}^{(n)} \boxplus \bm{\xi}^{(n)},
\end{equation}
where $ \Delta \bm{\xi} \boxplus \bm{T} = exp_{\mathfrak{se}(3)}(\Delta\bm{\xi}) \bm{T} $.  We also use a constant motion assumption, where the pose initialization is taken to be the relative pose motion from the previous update, as opposed to initializing with the identity pose.  The pose initialization for frame $F_i$ with respect to frame $F_k$ thus can be expressed as
\begin{equation}\label{eq:7}
    \bm{\xi}_{ki,init} = \bm{\xi}_{k,i-1} \boxplus \bm{\xi}_{i-2,i-1} .
\end{equation}.
Experimentally we have found that this greatly improves performance by providing the system with an accurate initialization such that it can converge to a low-error solution.

\begin{figure*}[t]
\begin{center}
\includegraphics[width=1.0\linewidth]{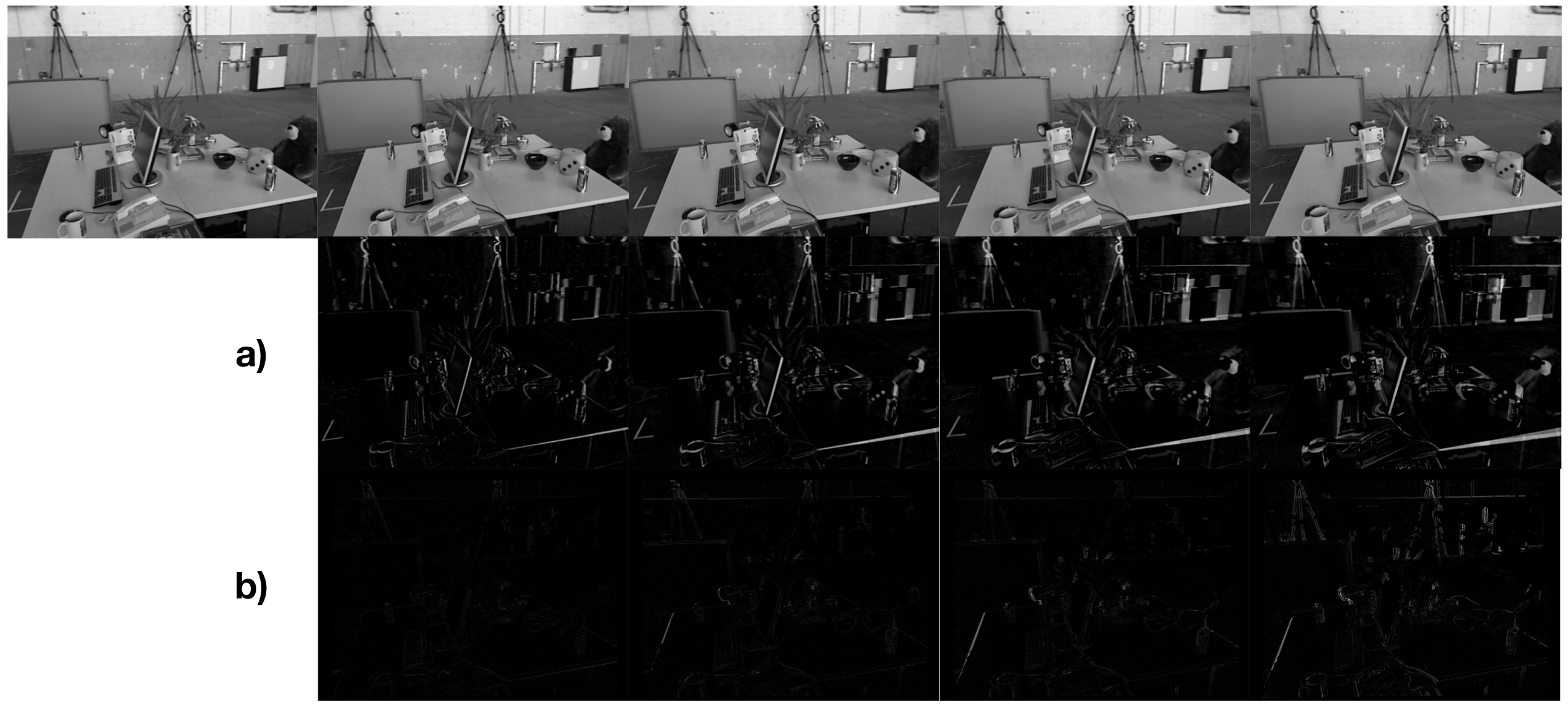}
\end{center}
   \caption{In the top row are the original images being localized with respect to the first image.  a) All original residuals b) All residuals after only minimizing edge residuals. This shows that minimizing the residuals for just the edge pixels jointly minimizes the residuals for \textit{all} pixels.  After 3 images, the minimization starts to become more inaccurate.  This is also a function of camera velocity and rotational velocity.}
\label{fig:residuals}
\end{figure*}

\subsection{Robust Gauss-Newton on Edge Maps}
Similar to other direct methods, we employ a coarse-to-fine approach to Gauss-Newton minimization to avoid false convergence.  The selection of the image pyramid scheme has a large effect on the system performance, and must be chosen carefully.  Some systems such as~\cite{engel14eccv} report using up to six levels, while~\cite{RobustKerl} report using four levels.  Simply extending these large pyramid sizes to edge maps causes the system to fail to converge to the correct solution.  This is due to the effects of aliasing.  A much smaller pyramid size is required. 

We found that three levels worked well for the original $640 \times 480$ resolution.  Using additional levels caused the system to fail due to edge aliasing effects which is illustrated in Figure~\ref{fig:aliasing}, which shows the same edge image at different levels of the pyramid.  After level three, it becomes unrecognizable.  For this reason, we recommend using images no smaller than $160 \times 120$ in resolution. 

A common approach in direct methods is to incorporate a weighting function that increases robustness to outliers when solving the error function. We use an iteratively re-weighted residual error function that we minimize with Gauss-Newton.  We found that iteratively re-weighting using Huber weights worked quite well for our application, following the work of~\cite{engel14eccv}.  The Huber weights are defined as
\begin{equation}\label{eq:8}
    w_i(r_i) = \begin{cases}
    1 , \quad r_i \leq k \\
    \frac{k}{|r_i|} , \quad r_i > k
    \end{cases} .
\end{equation}

The error function now becomes

\begin{equation}\label{eq:9}
     E(\bm{\xi}) = \sum_{i} w_i (\bm{\xi}) r_i^2 (\bm{\xi}) .
\end{equation}

Our goal is to find the relative camera pose that minimizes this function
\begin{equation}\label{eq:10}
    \arg\min_{\bm{\xi}} E(\bm{\xi}) = \arg\min_{\bm{\xi}} \sum_{i} w_i (\bm{\xi}) r_i^2 (\bm{\xi})  .
\end{equation}

In order to minimize this nonlinear error function with Gauss-Newton, we must linearize the equation. We can then solve this as a first-order approximation by iteratively solving the equation
\begin{equation}\label{eq:11}
    \Delta \xi^{(n)} = -(\bm{J}^T\bm{W}\bm{J})^{-1} \bm{J}^T \bm{W} r(\bm{\xi}^{(n)}) ,
\end{equation}
where $\bm{W}$ is a diagonal matrix with the weights, and the Jacobian $\bm{J}$ is defined as
\begin{equation}\label{eq:12}
    \bm{J} = \nabla \mathcal{I}_2 \frac{\partial \pi}{\partial \bm{P} } \frac{\partial \bm{P}}{\partial \bm{T}}\frac{\partial \bm{T}}{\partial \bm{\xi}} ,
\end{equation}
and $\nabla \mathcal{I}_2$ is the image gradient of the new image.  We can then iteratively update the relative pose with Equation \ref{eq:6}.

Note that we use the inverse-composition~\cite{LKframe} formulation such that we do not have to recompute the Jacobian matrix every iteration.  This is what makes this algorithm extremely efficient, as shown in~\cite{LKframe}.


\subsection{Optimizing over Edge Points}
We present the theory of selecting and incorporating edge points in the formulation, and provide some insight on why it is so effective in implementation. For edge selection process, note that we have two images, a reference and a new image, and therefore two sets of edges.  We wish to avoid the problems that arise from using both sets, namely there will be a different number of edge pixels, and dealing with this through matching algorithms is inefficient and error-prone. We use a more elegant solution, which is to use the edges of the new image as a mask on the first image. 

This initialization causes the mask to select pixels in the reference image that are slightly off from the reference image edges, assuming the camera has moved.  At each iteration, we follow a gradient from this position towards a point that reduces photometric error.  By definition, edges are regions of large photometric variation on either side.  Intuitively we argue that the optimization should therefore converge and settle at the correct edge.  To summarize, we initialize the edge mask to an offset position from the reference image's edges, and iteratively force these edge pixels to overlap with the reference edges.  In doing this we achieve a highly accurate relative pose.

\subsection{Keyframe Selection}

Another implementation detail has to do with keyframes.  Frame-to-frame alignment is inherently noisy and prone to accumulate drift.  To mitigate this, VO algorithms often select a key-frame which is used as the reference image for multiple new frames.  The error accumulation is  decreased by comparing against fewer reference frames, which directly results in a smaller error stackup.  

There have been several strategies for selecting keyframes.  The selection of keyframes is dependent on the type of VO algorithm being used.  Feature-based methods such as~\cite{orbslam} usually impose the restriction that a significant number of frames to pass, on the order of tens of frames.  

In~\cite{densergbdslam} the authors summarize several common approaches that direct methods use for creating a new keyframe: every $n$ frames, after a certain relative pose threshold has been met, the variance of the error function exceeds a threshold, or the differential entropy of the covariance matrix reaches a threshold.  However, each metric is not without its problems. 

Furthermore, the performance of the tracking degrades the further apart the baselines.  Figure~\ref{fig:residuals} demonstrates this phenomena, in which the residuals from five consecutive frames with respect to the first frame are shown. We observe that in general after 4 frames, the residuals become harder to minimize for most sequences.  Note that this is a function of camera motion.  We make the assumption that this camera tracking will be used for moderate motion and select an every $n$ frames approach.


\begin{table*}[t]
\begin{center}
\resizebox{\textwidth}{!}{\begin{tabular}{cccccccccccc}
\toprule
\multicolumn{11}{c}{Relative Pose Error (RPE) [m/s]} \\
 \midrule
 & ours & ours & ours & ours & ours & REVO\cite{robustedges} & REVO\cite{robustedges} & DSLAM\cite{densergbdslam} & ORB2\cite{orbslam2} & SLAM\cite{3dmappingslam} & D-EA\cite{deaedges} \\
  Seq. & Canny$_{KF}$ &Canny$_{FF}$ & LoG$_{FF}$  & Sobel$_{FF}$ & SE$_{FF}$ & SE$_{FF}$ & SE$_{KF}$ & ICP+Gray & Feat. & Feat.  & Canny \\
\midrule
fr1/xyz &0.02228 &0.02768 & 0.02821 & 0.02712 & 0.03289 & 0.03202 & 0.01957 & 0.02661 & \textbf{\textcolor{blue}{0.01470}} & 0.04193 & 0.04942 \\
fr1/rpy & - & 0.03126 & \textbf{\textcolor{blue}{0.02714}} & 0.03694 & 0.03041 &  0.03553 & 0.04037 & 0.04865 & 0.03221 & 0.07028 & 0.16150  \\
fr1/desk & \textbf{0.02664} & \textcolor{blue}{0.03022} & \textcolor{blue}{0.03022} & 0.03598 & 0.03056 & 0.07800 & 0.22196 & 0.04429 & 0.06178 & 0.05346 & 0.10654 \\
fr1/desk2 & - & \textbf{\textcolor{blue}{0.04387}} & 0.04953 & 0.05566 & 0.04490  & 0.07056 & 0.06703 &0.05722 & 0.06535 & 0.06955 & 0.20117   \\
fr1/room & -& 0.04830 & 0.06239 & 0.05240 & 0.05006 & \textcolor{blue}{0.04816} & \textbf{0.04272} &0.06427 & 0.07081 & 0.06666 & 0.21649 \\
fr1/plant & -& \textbf{\textcolor{blue}{0.02736}} & 0.02752 & 0.04171 & 0.05006 & 0.03063 & 0.02381 & 0.04362 & 0.04218 & 0.03789 &  0.34099  \\
fr2/desk & \textbf{0.01237} & \textcolor{blue}{0.01375} & \textcolor{blue}{0.01375} & 0.01800 & 0.03021  & 0.01426 & 0.02453 & 0.03248 & 0.03067 & 0.01400 &  0.09968 \\

\midrule
\multicolumn{11}{c}{Absolute Trajectory Error (ATE) [m]} \\
\midrule

fr1/xyz & 0.04567 & 0.04478 & 0.04461 & 0.05260 & 0.04115  & 0.09011 & 0.05375 &0.05760 & \textbf{\textcolor{blue}{0.00882}} & 0.01347 & 0.13006 \\
fr1/rpy & - & 0.05561 & \textcolor{blue}{0.03982} & 0.04795 & 0.04983 & 0.08933 & 0.07684 & 0.16341 & 0.08090 & \textbf{0.02874} & 0.14822  \\
fr1/desk & 0.03133 & 0.05387& 0.05358 & 0.05977 & \textcolor{blue}{0.04802} & 0.18648 & 0.54789 & 0.18251 & 0.09091& \textbf{0.02583} & 0.16376 \\
fr1/desk2 & - & 0.06798& 0.08384 & 0.08189 & \textcolor{blue}{0.06271} & 0.16866 & 0.18163 & 0.18861 & 0.10090 & \textbf{0.04256} & 0.44886  \\
fr1/room & - & 0.27382 & 0.34505 & 0.31586 & 0.34167 & 0.30594 & 0.28897 &0.21559 & \textcolor{blue}{0.20282} & \textbf{0.10117} & 0.60361  \\
fr1/plant & -& 0.07559 & 0.06708 & 0.09975 & \textcolor{blue}{0.06560} & 0.07300 & \textbf{0.05623} &0.12216 & 0.07234 & 0.06388 & 0.56927  \\
fr2/desk & 0.12540 & \textcolor{blue}{0.16664}& 0.18830 & 0.19074 & 0.27464 & 0.32902 & 0.09590 &0.46796 & 0.38657 & \textbf{0.09505} &  0.94546 \\
\bottomrule
\end{tabular}}
\end{center}
\caption{Comparison of the performance of our system using three different types of edges. \textcolor{blue}{Blue} denotes best performing frame-to-frame VO, excluding SLAM or keyframe systems. \textbf{Bold} denotes best performing system overall. A dashed line indicates that using keyframes did not improve performance.}
\label{tab:mainresults}
\end{table*}

\section{Experiments}
We evaluate our system using the TUM RGB-D benchmark~\cite{sturm12iros} , which is provided by the Technical University of Munich.  The benchmark has been widely used by various SLAM and VO algorithms to benchmark their accuracy and performance over various test sequences.  Each sequence contains RGB images, depth images, accelerometer data, as well as groundtruth.  The camera intrinsics are also provided.  Groundtruth was obtained by an external motion capture system through triangulation, and the data was synchronized.

There are several challenging datasets within this benchmark.  Each sequence ranges in duration, trajectory, and translational and rotational velocities.  We follow the work of~\cite{robustedges} which uses seven  sequences to benchmark their system performance so to achieve a direct comparison with other methods.  

\begin{figure}[t]
\begin{center}
 \includegraphics[width=1.0\linewidth]{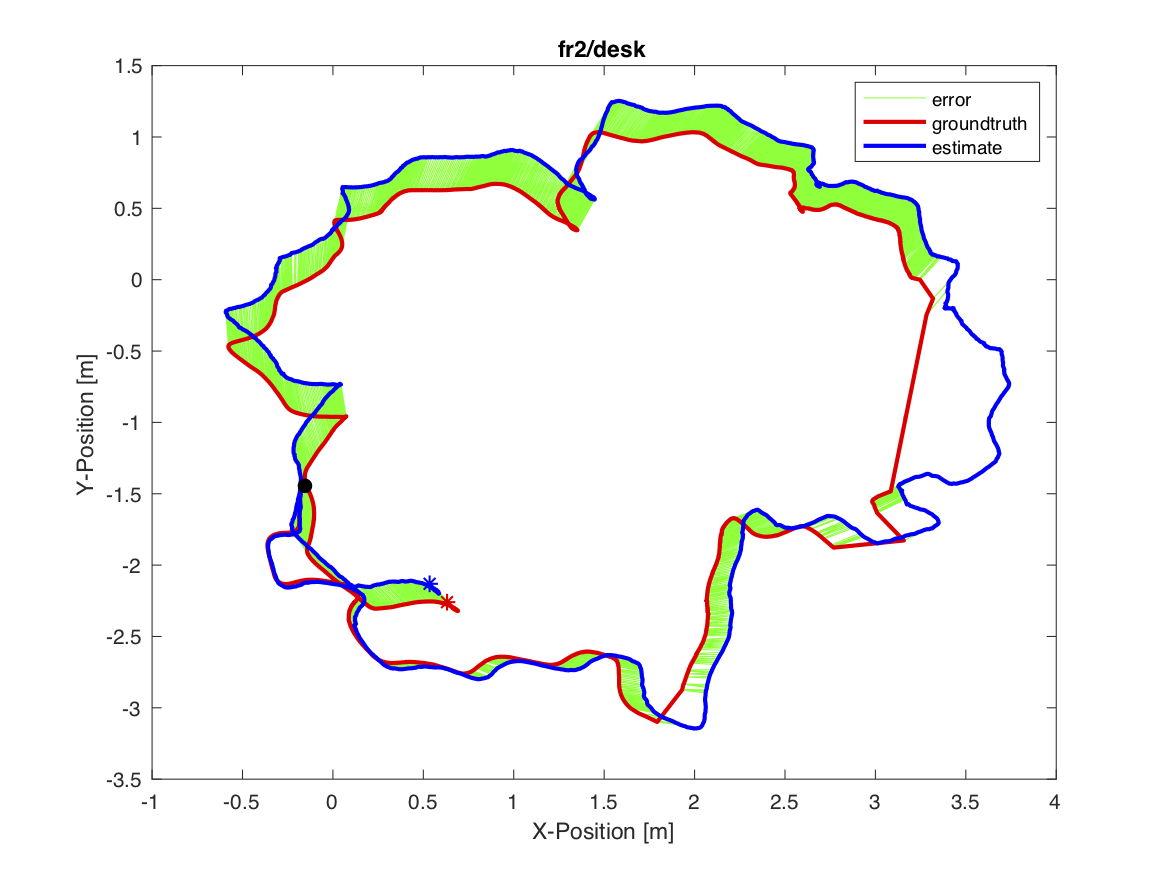}
\end{center}
   \caption{XY cross-section of our estimated trajectory compared with ground truth.  The error is shown in green.  The start position is shown as a black dot, while the final positions are shown as colored dots corresponding to the trajectory.  Areas without green indicate missing groundtruth data from sequence.  }
\label{fig:trajectory}
\end{figure}
\subsection{Evaluation Metrics}
We use the Relative Pose Error (RPE) and Absolute Trajectory Error (ATE) to evaluate our system.  The Relative Pose Error is proposed for evaluation of drift for VO algorithms in~\cite{sturm12iros}.  It measures the accuracy of the camera pose over a fixed time interval $\Delta t$ 

\begin{equation}\label{eq:rpe}
    RPE_t = (Q_t^{-1} Q_{t+\Delta t})(P_t^{-1} P_{t+\Delta t}),
\end{equation}

where $Q_1 \hdots Q_n \in SE(3)$ are the camera poses associated with the groundtruth trajectory and $P_1 \hdots P_n \in SE(3)$ are the camera poses associated with the estimated camera trajectory.  Similarly the Absolute Trjectory Error is defined as

\begin{equation}\label{eq:ate}
    ATE_t = Q_i^{-1} S P_i ,
\end{equation}
where poses $Q$ and $P$ are aligned by the rigid body transformation $S$ obtained through a least-squares solution.

A common practice has been to use the RMSE value of both the RPE and ATE, as RMSE values are a more robust metric that gives more weight to outliers as compared with the mean or median values. Thus the RMSE is a much more stringent performance metric to benchmark system drift.

Following the example set by \cite{Engel:semidense, RobustKerl, realtimeedgebased}, we provide the RMSE camera pose drift over several sequences of the dataset.  As first pointed out in~\cite{RobustKerl}, choosing too small of a $\Delta t$ creates erroneous error estimates as the ground truth motion capture system has finite error as well.  Too large of a value leads to penalizing rotations more so at the beginning than rotations towards the end~\cite{sturm12iros}.  Therefore, a reasonably sized $\Delta t$ needs to be chosen.  We use a $\Delta t$ of 1s to achieve direct comparison with other methods.  


\subsection{Results on the TUM RGB-D Benchmark}
We compare the performance of our algorithm using four different edge extraction algorithms, namely Canny, LoG, Sobel, and Structured Edges.  We compare to other methods using frame-to-frame tracking for all variants.  We selected Canny to perform keyframe tracking due to its consistent accuracy.  Although all of the edge types performed well on the sequences, Canny edges performed the best overall on average.  Note that we used automatic thresholding as opposed to REVO~\cite{robustedges} which used fixed threshold values, which introduces a dependency on photometric consistency.  Since we utilize automatic thresholding, our system is more robust to photometric variations across frames.  See Figure~\ref{fig:edgecomp} for examples of edge extractions.  

From our experiments we observed that edge-direct VO is highly accurate in frame-to-frame tracking, despite the inherent accumulation of drift in such a scheme that does not utilize keyframes.  In terms of RPE, our frame-to-frame variants perform better than or in worst case as well as REVO, an edge-based method which uses the distance transform on edges. Our method also outperforms ORB-SLAM2 run in VO mode for all sequences, except on $fr1/xyz$.  This is a result of ORB-SLAM2 keeping a local map, and in this particular sequence the camera keeps the majority of the initial scene in view at all times.  We confirmed this hypothesis by turning off the local mapping, at which case we outperform it on this sequence as well.   Our results are shown in Table~\ref{tab:mainresults}.  In terms of ATE, we again perform well across all non-SLAM algorithms.  Even though we do not use any Bundle Adjustment or global optimization as employed by RGBD-SLAM~\cite{densergbdslam}, we perform competitively over all sequences with such systems.

We provide plots of the edge-direct estimated trajectories over time compared to groundtruth in Figure~\ref{fig:sequence}.  Our estimated trajectory closely follows that of the groundtruth.  In Figure~\ref{fig:trajectory} we show the edge-direct estimated trajectory along the XY plane, along with the error between our estimate and groundtruth.

\subsection{Ablation Study}
In order to experimentally demonstrate the effect of using edge pixels we perform an ablation study. This two-fold ablation study demonstrates the relative efficacy between optimizing over edge pixels compared with optimizing over the same number of randomly chosen pixels, and additionally demonstrates the stability of using edge pixels.  We randomly select a fraction of the edge pixels to use, and compare it to our system randomly selecting the same number of pixels from the entire image.  We average over 5 runs to account for variability.  All parameters are identical for both methods. Additionally, for these tests we utilize keyframes as well as dropping the constant motion assumption.  This forces the system to rely on the optimization more heavily, and provides a better measurement of the quality of convergence. We additionally record the latency of our system per frame.  Operating on edge pixels is more accurate, while additionally enabling $\sim$50 fps on average on an Intel i7 CPU.  Note that at our optimization settings, a dense method is far from real-time.  

\begin{figure}[t]
\begin{center}
\includegraphics[width=1.0\linewidth]{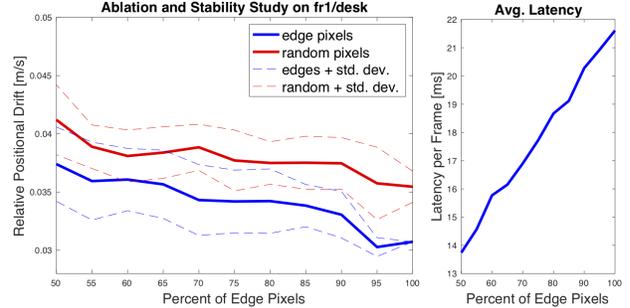}
\end{center}
   \caption{Left: Ablation study. Right: Frame-to-frame latency using edges.}
\label{fig:ablation}
\end{figure}

Since we use the Lucas-Kanade Inverse Compositional formulation we expected our algorithm to be linear time complexity with the number of pixels used.  We confirm this experimentally as well.  Refer to Figure~\ref{fig:ablation} for both the ablation study and timing measurements.  We save approximately 90\% computation on average by using edge pixels compared to using all pixels.  Note that for stability of edge pixels, the Kinect sensor used in the sequences filters out unstable points in its depth map, and from qualitative inspection still leaves a large number of reliable edge pixels.  This is confirmed via the relative stability of selected edge pixels compared to all pixels as well.  This ablation study further supports our claim that edge pixels are essential for robust and accurate camera tracking.


\section{Discussion}

\begin{figure*}[t]
\begin{center}
 \includegraphics[width=1.0\linewidth]{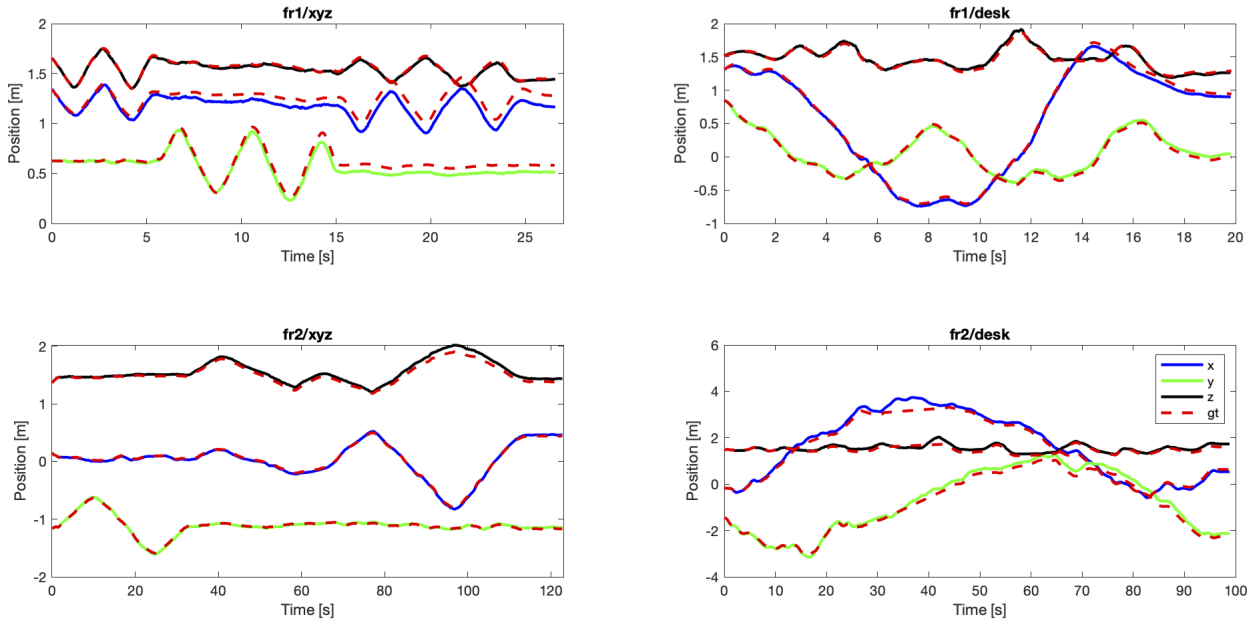}
\end{center}
   \caption{Shown is our estimated trajectory for four sequences.  Each sequence plots the trajectory in solid colors corresponding to the axis. Groundtruth is shown as a red dotted line for all axes.  As can be seen our estimates closely match that of the ground truth.  Note that for the sequence fr2/desk, there is no ground truth during the interval at approximately 31-43 seconds, which is why there appears to be a straight line in groundtruth trajectory.}
\label{fig:sequence}
\end{figure*}


Our edge-direct VO algorithm performs well across all sequences compared to other state-of-the-art methods. The trajectory in Figure~\ref{fig:trajectory} shows accurate camera tracking in a sequence that is 99 seconds long, and travels over 18 m without the use of Bundle Adjustment or loop closure. Note that our algorithm would perform even better if coupled with such global optimization methods, as our VO would initialize the algorithms closer to the correct solution compared with other algorithms.  Such an increase in accuracy can enable SLAM systems to rely less heavily on computationally expensive global optimizations, and perhaps run these threads less frequently.  Note that in this figure, the regions that are missing green regions are due to missing groundtruth data in the sequence. The estimated trajectory over time in Figure~\ref{fig:sequence} shows remarkably accurate results as well.  

It is important to note that even though we explicitly only minimize the photometric error for edge pixels, Figure~\ref{fig:residuals} shows that we simultaneously minimize the residuals for all pixels.  This is an important observation, as it supports the claim that minimizing the residuals of edge pixels is the minimally sufficient objective.  Moreover, the ablation study supports the claim that minimizing the photometric residuals for just the edge pixels provides less pixels to iterate over while enabling accurate tracking.

It is interesting to note that utilizing keyframes did not help the system improve on many of the sequences once we added the constant motion assumption.  Prior to adding this camera motion model, utilizing keyframes helped significantly.

\section{Conclusion}
We have presented a novel edge-direct visual odometry algorithm that determines an accurate relative pose between two frames by minimizing the photometric error of only edge pixels.  We demonstrate experimentally that minimizing the edge residuals jointly minimizes the residuals over the entire image.  This minimalist representation reduces computation required by operating on all pixels, and also results in more accurate tracking.  We benchmark its performance on the TUM RGB-D dataset where it achieves state-of-the-art performance as quantified by low relative pose drift and low absolute trajectory error.

\newpage
{\small
\bibliographystyle{ieee}

}

\end{document}